\pdfoutput=1

\documentclass[11pt]{article}

\usepackage[]{ACL2023}

\usepackage{times}
\usepackage{latexsym}

\usepackage[T1]{fontenc}

\usepackage[utf8]{inputenc}

\usepackage{microtype}

\usepackage{inconsolata}

\usepackage{graphicx}
\usepackage{ifthen}
\usepackage{xargs}
\usepackage{amsmath, amsthm, amssymb, amsfonts}
\usepackage{booktabs}

\newcommandx{\yaHelper}[2][1=\empty]{%
\ifthenelse{\equal{#1}{\empty}}%
  { \ensuremath{ \scriptstyle{ #2 } } } 
  { \raisebox{ #1 }[0pt][0pt]{ \ensuremath{ \scriptstyle{ #2 } } } }  
}   

\newcommandx{\yrightarrow}[4][1=\empty, 2=\empty, 4=\empty, usedefault=@]{%
  \ifthenelse{\equal{#2}{\empty}}
  { \xrightarrow{ \protect{ \yaHelper[ #4 ]{ #3 } } } } 
  { \xrightarrow[ \protect{ \yaHelper[ #2 ]{ #1 } } ]{ \protect{ \yaHelper[ #4 ]{ #3 } } } } 
}

\newcommandx{\yleftarrow}[4][1=\empty, 2=\empty, 4=\empty, usedefault=@]{%
  \ifthenelse{\equal{#2}{\empty}}
  { \xleftarrow{ \protect{ \yaHelper[ #4 ]{ #3 } } } } 
  { \xleftarrow[ \protect{ \yaHelper[ #2 ]{ #1 } } ]{ \protect{ \yaHelper[ #4 ]{ #3 } } } } 
}

\title{KGLM: Integrating Knowledge Graph Structure\\in Language Models for Link Prediction}

\author{
    Jason Youn\textsuperscript{\rm 1,2,3} \and Ilias Tagkopoulos \textsuperscript{\rm 1,2,3} \\
    \textsuperscript{\rm 1} Department of Computer Science, University of California, Davis, CA 95616, USA. \\ 
    \textsuperscript{\rm 2} Genome Center, University of California, Davis, CA 95616, USA. \\
    \textsuperscript{\rm 3} USDA/NSF AI Institute for Next Generation Food Systems (AIFS),\\University of California, Davis, CA 95616, USA. \\
    \{jyoun, itagkopoulos\}@ucdavis.edu
}

\begin{document}
\maketitle
\begin{abstract}
The ability of knowledge graphs to represent complex relationships at scale has led to their adoption for various needs including knowledge representation, question-answering, and recommendation systems. Knowledge graphs are often incomplete in the information they represent, necessitating the need for knowledge graph completion tasks. Pre-trained and fine-tuned language models have shown promise in these tasks although these models ignore the intrinsic information encoded in the knowledge graph, namely the entity and relation types. In this work, we propose the Knowledge Graph Language Model (KGLM) architecture, where we introduce a new entity/relation embedding layer that learns to differentiate distinctive entity and relation types, therefore allowing the model to learn the structure of the knowledge graph. In this work, we show that further pre-training the language models with this additional embedding layer using the triples extracted from the knowledge graph, followed by the standard fine-tuning phase sets a new state-of-the-art performance for the link prediction task on the benchmark datasets.
\end{abstract}

\section{Introduction}

\begin{figure}[t]
\centering
\includegraphics[width=1.0\columnwidth]{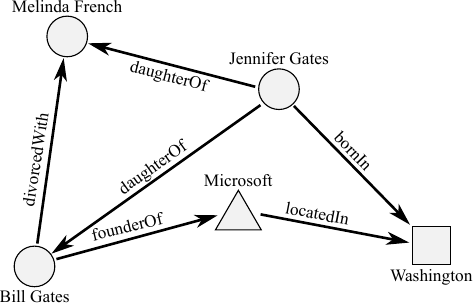}
\caption{Sample knowledge graph with 6 triples. The graph contains three unique entity types (circle for person, triangle for company, and square for location) and 5 unique relation types or 10 if considering both the forward and inverse relations. The task of the knowledge graph completion is to complete the missing links in the graph, e.g., (\emph{Bill Gates}, \emph{bornIn?}, \emph{Washington}) using the existing knowledge graph.}
\label{fig:sample_kg}
\end{figure}

Knowledge graph (KG) is defined as a directed, multi-relational graph where entities (nodes) are connected with one or more relations (edges) (\citealp{wang2017knowledge}). It is represented with a set of triples, where a triple consists of (\emph{head entity}, \emph{relation}, \emph{tail entity}) or ($h$, $r$, $t$) for short, for example (\emph{Bill Gates}, \emph{founderOf}, \emph{Microsoft}) as shown in Figure \ref{fig:sample_kg}. Due to their effectiveness in identifying patterns among data and gaining insights into the mechanisms of action, associations, and testable hypotheses (\citealp{li2014big, silvescu2012graph}), both manually curated KGs like DBpedia (\citealp{auer2007dbpedia}), WordNet (\citealp{miller1998wordnet}), KIDS (\citealp{youn2022knowledge}), and CARD (\citealp{alcock2020card}), and automatically curated ones like FreeBase (\citealp{bollacker2008freebase}), Knowledge Vault (\citealp{dong2014knowledge}), and NELL (\citealp{carlson2010toward}) exist. However, these KGs often suffer from incompleteness. For example, 71\% of the people in FreeBase have no known place of birth (\citealp{west2014knowledge}). To address this issue, knowledge graph completion (KGC) methods aim at connecting the missing links in the KG.

\begin{figure*}[ht]
\centering
\includegraphics[width=2\columnwidth]{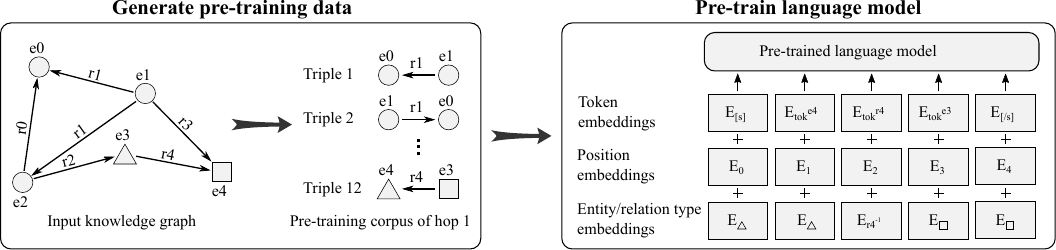}
\caption{Proposed pre-training approach of the KGLM. First, both the forward and inverse triples are extracted from the knowledge graph to serve as the pre-training corpus. We then continue pre-training the language model, RoBERTa in our case, using the masked language model training objective, with an additional entity/relation-type embedding layer. The entity/relation-type embedding scheme shown here corresponds to the KGLM\textsubscript{GER}, the most fine-grained version where both the entity and relation types are considered unique. Note that the inverse relation denoted by \textsuperscript{-1} is different from its forward counterpart. For demonstration purposes, we assume all entities and relations to have a single token.}
\label{fig:overall}
\end{figure*}

Graph feature models like path ranking algorithm (PRA) (\citealp{lao2010relational, lao2011random}) attempt to solve the KGC tasks by extracting the features from the observed edges over the KG to predict the existence of a new edge (\citealp{nickel2015review}). For example, the existence of the path \emph{Jennifer Gates} $\yrightarrow{\text{\emph{daughterOf}}}[-2pt]$ \emph{Melinda French} $\yleftarrow{\text{\emph{divorcedWith}}}[-2pt]$ \emph{Bill Gates} in Figure \ref{fig:sample_kg} can be used as a clue to infer the triple (\emph{Jennifer Gates}, \emph{daughterOf}, \emph{Bill Gates}). Other popular types of models are latent feature models such as TransE (\citealp{bordes2013translating}), TransH (\citealp{wang2014knowledge}), and RotatE (\citealp{sun2019rotate}) where entities and relations are converted into a latent space using embeddings. TransE, a representative latent feature model, models the relationship between the entities by interpreting them as a translational operation. That is, the model optimizes the embeddings by enforcing the vector operation of head entity embedding \emph{\textbf{h}} plus the relation embedding \emph{\textbf{r}} to be close to the tail entity embedding \emph{\textbf{t}} for a given fact in the KG, or simply $\emph{\textbf{h}} + \emph{\textbf{r}} \approx \emph{\textbf{t}}$.

Recently, pre-trained language models like BERT (\citealp{devlin2018bert}) and RoBERTa (\citealp{liu2019roberta}) have shown state-of-the-art performance in all of the natural language processing (NLP) tasks. As a natural extension, models like KG-BERT (\citealp{yao2019kg}) and BERTRL (\citealp{zha2021inductive}) that utilize these pre-trained language models by treating a triple in the KG as a textual sequence, e.g., (\emph{Bill Gates}, \emph{founderOf}, \emph{Microsoft}) as `\emph{Bill Gates founder of Microsoft}', have also shown state-of-the-art results on the downstream KGC tasks. Although such \emph{textual encoding} (\citealp{wang2021structure}) models are generalizable to unseen entities or relations (\citealp{zha2021inductive}), they still fail to learn the intrinsic structure of the KG as the models are only trained on the textual sequence. To solve this issue, a hybrid approach like StAR (\citealp{wang2021structure}) has recently been proposed to take advantage of both latent feature models and textual encoding models by enforcing a translation-based graph embedding approach to train the textual encoders. Yet, current textual encoding models still suffer from entity ambiguation problems (\citealp{cucerzan2007large}) where an entity \emph{Apple}, for example, can refer to either the company Apple Inc. or the fruit. Moreover, there are no ways to distinguish forward relation (\emph{Jennifer Gates}, \emph{daughterOf}, \emph{Melinda French}) from inverse relation (\emph{Melinda French}, \emph{daughterOf\textsuperscript{  -1}}, \emph{Jennifer Gates}).

In this paper, we propose the Knowledge Graph Language Model (KGLM) (Figure \ref{fig:overall}), a simple yet effective language model pre-training approach that learns from both the textual and structural information of the knowledge graph. We continue pre-training the language model that has already been pre-trained on other large natural language corpora using the corpus generated by converting the triples in the knowledge graphs as textual sequences, while enforcing the model to better understand the underlying graph structure and by adding an additional entity/relation-type embedding layer. Testing our model on the WN18RR dataset for the link prediction task shows that our model improved the mean rank by 21.2\% compared to the previous state-of-the-art method (51 vs. 40.18, respectively). All code and instructions on how to reproduce the results are available online.\footnote{\url{https://github.com/ibpa/KGLM}}

\section{Background}

\textbf{Link Prediction.} The link prediction (LP) task, one of the commonly researched knowledge graph completion tasks, attempts to predict the missing head entity ($h$) or tail entity ($t$) of a triple ($h$, $r$, $t$) given a KG $G=(E,R)$, where $\{h, t\} \in E$ is the set of all entities and $r \in R$ is the set of all relations. Specifically, given a single test positive triple ($h$, $r$, $t$), its corresponding link prediction test dataset can be constructed by corrupting either the head or the tail entity in the filtered setting (\citealp{bordes2013translating}) as

\begin{equation}
\begin{gathered}
\mathcal{D}_{LP}^{(h,r,t)} = \\
\{ (\emph{h}, \emph{r}, \emph{t'}) \; | \; t' \in (E-\{h,t\}) \wedge (h,r,t') \notin \mathcal{D} \} \\
\cup \{ (\emph{h'}, \emph{r}, \emph{t}) \; | \; h' \in (E-\{h,t\}) \wedge (h',r,t) \notin \mathcal{D} \} \\
\cup \{ (h,r,t) \},
\label{equ:link_prediction_dataset}
\end{gathered}
\end{equation}

where $\mathcal{D}=\mathcal{D}_{train} \cup \mathcal{D}_{val} \cup \mathcal{D}_{test}$ is the complete dataset. Evaluation of the link prediction task is measured with mean rank (MR), mean reciprocal rank (MRR), and hits@N (\citealp{rossi2021knowledge}). MR is defined as

\begin{equation}
MR = \frac{ \sum\limits_{(h,r,t) \in \mathcal{D}_{test}} rank((h,r,t) \; | \; \mathcal{D}_{LP}^{(h,r,t)}) }{|\mathcal{D}_{test}|},
\label{equ:MR}
\end{equation}

where $rank(\cdot|\cdot)$ is the rank of the positive triple among its corrupted versions and $|\mathcal{D}_{test}|$ is the number of positive test triples. MRR is the same as MR except that the reciprocal rank $1/rank(\cdot|\cdot)$ is used. Hits@N is defined as 

\begin{equation}
\begin{gathered}
hits@N = \\
\frac{ \sum\limits_{(h,r,t) \in \mathcal{D}_{test}} \begin{cases} 1, & \!\!\!\! $if $ rank((h,r,t) \; | \; \mathcal{D}_{LP}^{(h,r,t)})<N \\ 0, & \!\!\!\! otherwise \end{cases} }{|\mathcal{D}_{test}|},
\label{equ:hits@N}
\end{gathered}
\end{equation}

where $N \in \{1,3,10\}$ is commonly reported. Higher MRR and hits@N values are better, whereas, for MR, lower values denote higher performance.

\section{Proposed Approach}
In this work, we propose to continue pre-training, instead of pre-training from scratch, the language model RoBERTa\textsubscript{LARGE} (\citealp{liu2019roberta}) that has already been trained on English-language corpora of varying sizes and domains, using both the forward and inverse knowledge graph textual sequences (Figure \ref{fig:overall}). Following the convention used in the KG-BERT and StAR (see Appendix~\ref{sec:previous_work}), we use a textual representation of a given triple, e.g., (\emph{Bill Gates}, \emph{founderOf}, \emph{Microsoft}) as `\emph{Bill Gates founder of Microsoft}', to generate the pre-training corpus. However, instead of extracting only the forward triple as done in the previous work, we extract both the forward and inverse versions of the triple, e.g., (\emph{Jennifer Gates}, \emph{daughterOf}, \emph{Bill Gates}) and (\emph{Bill Gates}, \emph{daughterOf\textsuperscript{  -1}}, \emph{Jennifer Gates}), where the \textsuperscript{-1} notation denotes the inverse direction of the corresponding relation.

\setlength{\tabcolsep}{3.5pt}
\begin{table}[t]
 \caption{Statistics of the benchmark knowledge graphs used for link prediction.}
 \small
  \centering
  \begin{tabular}{cccccc}
    \toprule
    Dataset & \# ent & \# rel & \# train & \# val & \# test \\
    \midrule
    WN18RR & 40,943 & 11 & 86,835 & 3,034 & 3,134 \\
    FB15k-237 & 14,951 & 237 & 272,115 & 17,535 & 20,466 \\
    UMLS & 135 & 46 & 5,216 & 652 & 661 \\
    \bottomrule
  \end{tabular}
  \label{tab:datasets}
\end{table}

To enforce the model to learn the knowledge graph structure, we introduce a new embedding layer \emph{entity/relation-type embedding} (ER-type embedding) in addition to the pre-existing token and position embeddings of RoBERTa as shown in Figure \ref{fig:overall}. This additional layer aims to embed the tokens in the input sequence with its corresponding entity/relation-type, where the set of entities $E$ in the knowledge graph can have $t_{E}$ different entity types depending on the schema of the knowledge graph, (e.g., $t_{E}=3$ for person, company, and location in Figure \ref{fig:sample_kg}). Note that many knowledge graphs do not specify the entity types, in which case $t_{E} = 1$. For the set of relations $R$, there exist $t_{R}=2n_{R}$, where $n_{R}$ is the number of unique relations in the knowledge graph and the multiplier of 2 comes from forward and inverse directions (e.g., $t_{R}=10$ for the sample knowledge graph in Figure \ref{fig:sample_kg}).

\begin{table*}[ht]
\caption{Link prediction results on the benchmark datasets WN18RR, FB15k-237, and UMLS. Bold numbers denote the best performance for a given metric and class of models. Underlined numbers denote the best performance for a given metric regardless of the model type. Note that we do not report KGLM\textsubscript{GER} performance since the tested datasets do not specify entity types in their schema.}
\small
\centering
\setlength{\tabcolsep}{3.5pt}
\begin{tabular}{ccccccccccccc}
    \toprule
    \multicolumn{1}{c}{} & \multicolumn{5}{c}{WN18RR} & \multicolumn{5}{c}{FB15k-237} & \multicolumn{2}{c}{UMLS} \\
    \cmidrule(r){2-6} \cmidrule(r){7-11} \cmidrule(r){12-13}
    Method & Hits @1 & Hits @3 & Hits @10 & MR & MRR & Hits @1 & Hits @3 & Hits @10 & MR & MRR & Hits@10 & MR\\
    \midrule
    \multicolumn{6}{l}{\emph{Model type: Not based on language models}} \\
    \midrule
    TransE & .043 & .441 & .532 & 2300 & .243 & .198 & .376 & .441 & 323 & .279 & .989 & 1.84 \\
    TransH & .053 & .463 & .540 & 2126 & .279 & .306 & .450 & .613 & 219 & .320 & - & - \\
    DistMult & .412 & .470 & .504 & 7000 & .444 & .199 & .301 & .446 & 512 & .281 & .846 & 5.52 \\
    ComplEx & .409 & .469 & .530 & 7882 & .449 & .194 & .297 & .450 & 546 & .278 & .967 & 2.59\\
    ConvE & .390 & .430 & .480 & 5277 & .46 & .239 & .350 & .491 & 246 & .316 & \textbf{.990} & \textbf{1.51}\\
    RotatE & .428 & .492 & .571 & 3340 & .476 & .241 & .375 & .533 & 177 & .338 & - & -\\
    GAAT & .424 & \textbf{.525} & \textbf{.604} & \textbf{1270} & .467 & \underline{\textbf{.512}} & \underline{\textbf{.572}} & \underline{\textbf{.650}} & 187 & \underline{\textbf{.547}} & - & - \\
    LineaRE & \underline{\textbf{.453}} & .509 & .578 & 1644 & \underline{\textbf{.495}} & .264 & .391 & .545 & 155 & .357 & - & - \\
    QuatDE & .438 & .509 & .586 & 1977 & .489 & .268 & .400 & .563 & \underline{\textbf{90}} & .365 & - & - \\
    \midrule
    \multicolumn{6}{l}{\emph{Model type: Based on language models}} \\
    \midrule
    KG-BERT & .041 & .302 & .524 & 97 & .216 & - & - & .420 & 153 & - & .990 & 1.47 \\
    StAR & .243 & .491 & .709 & 51 & .401 & \textbf{.205} & \textbf{.322} & \textbf{.482} & \textbf{117} & \textbf{.296} & .991 & 1.49 \\
    \midrule
    \textbf{KGLM\textsubscript{Base}} & .305 & .518 & .730 & 47.97 & .445 & - & - & - & - & - & - & - \\
    \textbf{KGLM\textsubscript{GR}} & \textbf{.330} & \underline{\textbf{.538}} & \underline{\textbf{.741}} & \underline{\textbf{40.18}} & \textbf{.467} & .200 & .314 & .468 & 125.9 & .289 & \underline{\textbf{.995}} & \underline{\textbf{1.19}} \\
    \bottomrule
\end{tabular}
\label{tab:link_prediction_results}
\end{table*}

In this work, we propose three different variations of ER-type embeddings. KGLM\textsubscript{Base} is the simplified version where all entities are assigned a single entity type and relations are assigned either forward or inverse relation type regardless of their unique relation types, resulting in a total of 3 ER-type embeddings. The KGLM\textsubscript{GR} is a version with granular relation types with $t_{R} + 1$ ER-type embeddings. The KGLM\textsubscript{GER} is the most granular version where we utilize all $t_{E}+t_{R}$ ER-type embeddings. In other words, all entity types as well as all relation types including both directions are considered.

To be specific, we convert a triple ($h$, $r$, $t$) to a sequence of tokens $w^{(h,r,t)} = \langle \texttt{[s]} w_a^h w_b^r w_c^t \texttt{[/s]} : a\in\{1..|h|\} {\And} b\in\{1..|r|\} {\And} c\in\{1..|t|\} \rangle \in \mathbb{R}^{(|h|+|r|+|t|+2)}$, where \texttt{[s]} and \texttt{[/s]} are special tokens denoting beginning and end of the sequence, respectively. The input to the RoBERTa model is then constructed by adding the ER-type embedding $\mathbf{t}^{(h,r,t)}$ and the $\mathbf{p}^{(h,r,t)}$ position embeddings to the $\mathbf{w}^{(h,r,t)}$ token embeddings, as
\begin{equation}
 \mathbf{X}^{(h,r,t)} = \mathbf{w}^{(h,r,t)} + \mathbf{p}^{(h,r,t)} + \mathbf{t}^{(h,r,t)}.
\label{equ:input_pathlm}
\end{equation}
Unlike the segment embeddings in the KG-BERT and StAR that were used to mark the input tokens with either the entity ($\mathbf{s}_e$) or relation ($\mathbf{s}_r$), the ER-type embedding now replaces its functionality. Finally, we pre-train the model using the masked language model (MLM) training objective (\citealp{liu2019roberta}).

For fine-tuning, we extend the idea of how the KG-BERT scores a triple (see Equation \ref{equ:score_kgbert} in Appendix~\ref{sec:previous_work}) to take advantage of the ER-type embeddings learned in our pre-training stage. For a given target triple, we calculate the weighted average score of both directions as
\begin{equation}
\begin{gathered}
score_{KGLM}(h,r,t) = \alpha \text{SeqCls}(\mathbf{X}^{(h,r,t)}) + \\
(1 - \alpha) \text{SeqCls}(\mathbf{X}^{(t,r^{-1},h)}),
\label{equ:score_seq_cls_hop1}
\end{gathered}
\end{equation}
where SeqCls($\cdot$) is a RoBERTa model transformer with a sequence classification head on top of the pooled output (last layer hidden-state of the \texttt{[CLS]} token followed by dense layer and $\tanh$ activation function), ${(t,r^{-1},h)}$ denotes the inverse version of $(h,r,t)$, and $0 \leq \alpha \leq 1$ denotes the weight used for balancing the scores from forward and inverse scores. For example, $\alpha = 1.0$ considers only the forward direction score.

\section{Experiments and Results}

\subsection{Datasets}
We tested our proposed method on three benchmark datasets WN18RR, FB15k-237, and UMLS as shown in Table \ref{tab:datasets}. WN18RR (\citealp{dettmers2018convolutional}) is derived from WordNet (\citealp{miller1998wordnet}), a large English lexical database of semantic relationships between words, FB15k-237 (\citealp{toutanova2015observed}) is extracted from Freebase (\citealp{bollacker2008freebase}), a large community-drive KG of general facts about the world, and UMLS contains biomedical relationships. WN18RR and FB15k-237 are subsets of WN18 (\citealp{bordes2013translating}) and FB15k (\citealp{bordes2013translating}), respectively, where the \emph{inverse relation test leakage} problem, i.e. the problem of inverted test triples appearing in the training set, has been corrected.

\begin{table*}[ht]
 \caption{Breakdown of the original hypothesis and their results on WN18RR. For claim 1, we continued to pre-train RoBERTa\textsubscript{LARGE} using the knowledge graph without the ER-type embeddings. Note that we did not also use the ER-type embeddings layer in the fine-tuning stage. For claim 2, we learned the ER-type embeddings in the fine-tuning stage only without any further pre-training.}
 \small
  \centering
  \begin{tabular}{ccccccccc}
    \toprule
    \multicolumn{2}{c}{} & \multicolumn{2}{c}{ER-type embeddings} & \multicolumn{5}{c}{} \\
    \cmidrule(r){3-4}
    Model & Continue pre-training & Pre-train & Fine-tune & Hits @1 & Hits @3 & Hits @10 & MR & MRR \\
    \midrule
    Claim 1 & o & x & x & \textbf{0.331} & 0.529 & 0.728 & 53.5 & 0.462 \\
    Claim 2 & x & - & o & 0.322 & 0.489 & 0.672 & 66.4 & 0.439 \\
    KGLM\textsubscript{GR} & o & o & o & 0.330 & \textbf{0.538} & \textbf{0.741} & \textbf{40.18} & \textbf{0.467} \\
    \bottomrule
  \end{tabular}
  \label{tab:breakdown}
\end{table*}

\subsection{Settings}
We used RoBERTa\textsubscript{LARGE} (\citealp{liu2019roberta}), a BERT\textsubscript{LARGE}-based architecture with 24 layers, 1024 hidden size, 16 self-attention heads, and 355M parameters, for the pre-trained language model as it has been shown in a previous study to perform better than BERT (hits@1 0.243 vs. 0.222 and MR 51 vs. 99, link prediction on WN18RR) (\citealp{wang2021structure}). For pre-training, we used learning rate = 5e-05, batch size = 32, epoch = 20 (WN18RR), 10 (FB15k-237), and 1,000 (UMLS), and AdamW optimizer (\citealp{loshchilov2017decoupled}). For fine-tuning training data, we sampled 10 negative triples for a positive triple by corrupting both the head and tail entity 5 times each. We used the validation set to find the optimal learning rates = $\{1e-06, 5e-07\}$, batch size = $\{16, 32\}$, epochs = $\{1, 2, 3, 4, 5\}$ for WN18RR and FB15k-237 and {25, 50, 75, 100} for UMLS, and $\alpha$ from 0.0 to 1.0 with an increment of 0.1. For all experiments, we set $\alpha=0.5$ based on the WN18RR validation set performance. Both pre-training and fine-tuning were performed on 3 $\times$ Nvidia Quadro RTX 6000 GPUs in a distributed manner using the 16-bit mixed precision and DeepSpeed (\citealp{rasley2020deepspeed, rajbhandari2020zero}) library in the stage-2 setting. We used the Transformers library (\citealp{wolf2019huggingface}).

\subsection{Link Prediction Results}
The hypothesis behind the KGLM was that learning the ER-type embedding layers in the pre-training stage using the corpus generated by the knowledge graph, followed by fine-tuning has the best performance. To test our hypothesis, we broke down the hypothesis into two separate claims. For the first claim, we only continued pre-training RoBERTa\textsubscript{LARGE} followed by fine-tuning without the ER-type embeddings. This test removes the contribution from the ER-type embeddings and solely tests the performance gained by further pre-training the model with the knowledge graph as input. Table \ref{tab:breakdown} shows that claim 1 falls behind the KGLM\textsubscript{GR} in all metrics except for hits @1 (0.331 vs. 0.330, respectively). For the second claim, we did not continue pre-training and instead used the RoBERTa\textsubscript{LARGE} pre-trained weights as-is. We then learned the ER-type embeddings in the fine-tuning stage. This test shows if the ER-type embeddings can be learned only during the fine-tuning stage. Table \ref{tab:breakdown} shows that KGLM\textsubscript{GR} outperforms all of the metrics obtained using the second claim. This result shows that the combination of these two claims works in a non-linear fashion to maximize performance.

The results of performing link prediction on the benchmark datasets are shown in Table \ref{tab:link_prediction_results}. Compared to StAR, which had the best performance on MR and hits@10 on WN18RR, KGLM\textsubscript{GR} outperformed all the metrics with 21.2\% improved MR (40.18 vs. 51, respectively) and 4.5\% increased hits@10 (0.709 vs. 0.741, respectively). Although still inferior compared to the graph embedding approaches, KGLM\textsubscript{GR} has 35.8\% improved hits@1 compared to the best language model-based approach StAR (0.243 vs. 0.330, respectively). Across all model types, KGLM\textsubscript{GR} has the best performance on all metrics for WN18RR except for hits@1. Although we did not observe any improvement compared to StAR for the FB15k-237 dataset, we had the best performance on all metrics for UMLS with 21.2\% improved MR than ComplEx (1.19 vs. 1.51, respectively). KGLM\textsubscript{GR} outperformed KGLM\textsubscript{Base} in all metrics.

\section{Conclusion}
In this work, we presented KGLM, which introduces a new entity/relation (ER)-type embedding layer for learning the structure of the knowledge graph. Compared to the previous language model-based methods that only fine-tune for a given task, we found that learning the ER-type embeddings in the pre-training stage followed by fine-tuning resulted in better performance. In future work, we plan to further test the version of KGLM that takes into account entity types,  KGLM\textsubscript{{GER}}, on domain-specific knowledge graphs like KIDS (\citealp{youn2022knowledge}) with entity types in their schema.

\section*{Limitations}
Although KGLM outperforms state-of-the-art models when the training set includes full sentences (e.g., UMLS and WN18RR), the model performed similarly to the state-of-the-art in cases where the training dataset had only ontological relationships, such as the /music/artist/origin relation present in the FB15k-237 dataset. One major limitation of the proposed method is the long training and inference time, which we plan to alleviate by adopting Siamese-style textual encoders (\citealp{wang2021structure, li2022lp}) in future work.

\section*{Ethics Statement}
The authors declare no competing interests.

\section*{Acknowledgements}
We would like to thank the members of the Tagkopoulos lab for their suggestions. This work was supported by the USDA-NIFA AI Institute for Next Generation Food Systems (AIFS), USDA-NIFA award number 2020-67021-32855 and the NIEHS grant P42ES004699 to I.T. J.Y. conceived the project and performed all experiments. Both J.Y. and I.T. wrote the manuscript. I.T. supervised all aspects of the project.

\bibliography{anthology,custom}
\bibliographystyle{acl_natbib}

\appendix

\section{Previous Work}
\label{sec:previous_work}

\subsection{KG-BERT}
KG-BERT (\citealp{yao2019kg}) is a fine-tuning method that utilizes the base version of the pre-trained language model BERT (BERT\textsubscript{BASE}) (\citealp{devlin2018bert}) as an encoder for entities and relations of the knowledge graph. Specifically, KG-BERT first converts a triple ($h$, $r$, $t$) to a sequence of tokens $w^{(h,r,t)} = \langle \texttt{[CLS]} w_a^h \texttt{[SEP]} w_b^r \texttt{[SEP]} w_c^t \texttt{[SEP]} : a\in\{1..|h|\} {\And} b\in\{1..|r|\} {\And} c\in\{1..|t|\} \rangle$, where $w_n$ denotes the n\textsuperscript{th} token of either entity or relation, \texttt{[CLS]} and \texttt{[SEP]} are the special tokens, while $|h|$, $|r|$, and $|t|$ denote the number of tokens in the head entity, relation, and tail entity, respectively. This textual token sequence is then converted to a sequence of token embeddings $\mathbf{w}^{(h,r,t)} \in \mathbb{R}^{d\times(|h|+|r|+|t|+4)}$, where $d$ is the dimension of the embeddings and 4 is from the special tokens. Then the segment embeddings $\mathbf{s}^{(h,r,t)} = \langle (\mathbf{s}_e)_{\times(|h|+2)} (\mathbf{s}_r)_{\times(|r|+1)} (\mathbf{s}_e)_{\times(|t|+1)} \rangle$, where $\mathbf{s}_e$ and $\mathbf{s}_r$ are used to differentiate entities from relations, respectively, as well as the position embeddings $\mathbf{p}^{(h,r,t)} = \langle \mathbf{p}_i : i \in \{1..(|h|{+}|r|{+}|t|{+}4)\} \rangle$ are added to the token embeddings $\mathbf{w}^{(h,r,t)}$ to form a final input representation $\mathbf{X}^{(h,r,t)} \in \mathbb{R}^{d\times(|h|+|r|+|t|+4)}$ that is fed to BERT as input. Then, the score of how likely a given triple ($h$, $r$, $t$) is to be true is computed by

\begin{equation}
score_{\text{KG-BERT}}(h,r,t) = \text{SeqCls}(\mathbf{X}^{(h,r,t)}).
\label{equ:score_kgbert}
\end{equation}

KG-BERT significantly improved the MR of the link prediction task compared to the previous state-of-the-art approach CapsE (\citealp{vu2019capsule}) (97 compared to 719, an 86.5\% decrease), but suffered from poor hits@1 of 0.041 due to the entity ambiguation problem and lack of structural learning (\citealp{wang2021structure, cucerzan2007large}).

\subsection{StAR}
StAR (\citealp{wang2021structure}) is a hybrid model that learns both the contextual and structural information of the knowledge graph by augmenting the structured knowledge in the encoder. It divides a triple into two parts, (\emph{h}, \emph{r}) and (\emph{t}), and applies a Siamese-style transformer with a sequence classification head to generate $\boldsymbol{u} = \text{Pool}(\mathbf{X}^{(h,r)}) \in \mathbb{R}^{d\times(|h|+|r|+3)}$ and $\boldsymbol{v} = \text{Pool}(\mathbf{X}^{(t)}) \in \mathbb{R}^{d\times(|t|+2)}$, respectively, where Pool($\cdot$) is the output of the RoBERTa's pooling layer. The first scoring module focuses on classifying the triple by applying a

\begin{equation}
score_{\text{StAR}}^{c}(h,r,t) = \text{Cls}([\boldsymbol{u}; \boldsymbol{u}\times\boldsymbol{v};\boldsymbol{u} - \boldsymbol{u}; \boldsymbol{v}]),
\label{equ:score_star_1}
\end{equation}

where Cls($\cdot$) is a neural binary classifier with a dense layer followed by a softmax activation function. The second scoring module then adopts the idea of how translation-based graph embedding methods like TransE learns the graph structure by minimizing the distance between $\boldsymbol{u}$ and $\boldsymbol{v}$ as

\begin{equation}
score_{\text{StAR}}^{d}(h,r,t) = -||\boldsymbol{u}-\boldsymbol{v}||,
\label{equ:score_star_2}
\end{equation}

where $||\cdot||$ is the \emph{L2}-normalization. During the training, StAR uses a weighted average of the binary cross entropy loss computed using $score_{\text{StAR}}^{c}(h,r,t)$ and the margin-based hinge loss computed using $score_{\text{StAR}}^{d}(h,r,t)$, whereas only the $score_{\text{StAR}}^{c}(h,r,t)$ is used for inference. This approach shows a new state-of-the performance over the metrics MR (51) and hits@10 (0.709), as well as significantly improving the hits@1 compared to the KG-BERT (0.041 to 0.243, a 492.7\% increase).

\end{document}